\PassOptionsToPackage{bookmarksnumbered,unicode,pdfpagelabels=false}{hyperref}
\RequirePackage{hyperref}
\PassOptionsToPackage{nopatch=footnote}{microtype}
\documentclass[sigconf]{acmart}

\makeatletter
\let\sj@origwarn\@latex@warning@no@line
\def\@latex@warning@no@line#1{\in@{Text page}{#1}\ifin@\else\sj@origwarn{#1}\fi}
\makeatother
\usepackage{booktabs}
\usepackage{multirow}
\usepackage{makecell}
\usepackage{tabularx}
\usepackage{subcaption}
\usepackage{verbatim}
\usepackage{xurl}
\usepackage{tikz}
\usetikzlibrary{positioning,arrows.meta,fit,backgrounds,calc}
\usepackage{pgfplots}
\pgfplotsset{compat=1.17}
\copyrightyear{2026}
\acmYear{2026}
\setcopyright{cc}
\setcctype{by}
\acmConference[ICVGIP 2026]{Indian Conference on Computer Vision, Graphics, and Image Processing}{December 21--24, 2026}{Kolkata, India}
\acmBooktitle{Indian Conference on Computer Vision, Graphics, and Image Processing (ICVGIP 2026), December 21--24, 2026, Kolkata, India}
\acmPrice{}
\acmDOI{}
\acmISBN{}

\newcommand{\eg}{\textit{e.g.}}
\newcommand{\ie}{\textit{i.e.}}
\newcommand{\etal}{\textit{et al.}}
\newcommand{\modelname}{\textsc{StyleJudge}}

\begin{document}

\title{StyleJudge: A Human-Validated, In-Context-Learning Vision-Language Evaluator for Artistic Style Transfer}

\author{Lokesh}
\affiliation{\institution{Indian Institute of Technology Mandi}\city{Mandi}\country{India}}
\email{lokeshyadav71709@gmail.com}
\author{Pramukto Mandal}
\affiliation{\institution{Indian Institute of Technology Mandi}\city{Mandi}\country{India}}
\email{pramuktoofficial@gmail.com}
\author{Somanshu Juneja}
\affiliation{\institution{Indian Institute of Technology Mandi}\city{Mandi}\country{India}}
\email{junejasomanshu@gmail.com}
\author{Jashnoor Singh}
\affiliation{\institution{Indian Institute of Technology Mandi}\city{Mandi}\country{India}}
\email{jashnoors77@gmail.com}
\author{Amit Kumar}
\affiliation{\institution{Indian Institute of Technology Mandi}\city{Mandi}\country{India}}
\email{arunkumarbahilwara@gmail.com}
\author{Pranav Adlinge}
\affiliation{\institution{Samsung R\&D Institute India-Bangalore}\city{Bangalore}\country{India}}
\email{p.adlinge@samsung.com}
\author{Richa Thakur}
\affiliation{\institution{Indian Institute of Technology Mandi}\city{Mandi}\country{India}}
\email{d23151@students.iitmandi.ac.in}
\author{Kunal Swami}
\affiliation{\institution{Samsung R\&D Institute India-Bangalore}\city{Bangalore}\country{India}}
\email{kunal.swami@samsung.com}
\author{Aditya Nigam}
\affiliation{\institution{Indian Institute of Technology Mandi}\city{Mandi}\country{India}}
\email{aditya@iitmandi.ac.in}
\renewcommand{\shortauthors}{Lokesh et al.}

\begin{abstract}
Automatic evaluation of artistic style transfer remains an open problem: existing metrics
score an image pair in isolation, without ever seeing an example of the intended style
transformation, and correlate poorly with human judgment. We introduce \modelname{}, a
compact vision-language evaluator that is instead conditioned on a small set of reference
source-to-stylized pairs before it scores a new query pair, mirroring how a human rater
calibrates to a style before judging a candidate. On a human-annotated benchmark of 878 rated pairs over 12 artistic styles (728 pairs from the ten training and validation styles, 150 from two unseen styles), \modelname{}, an efficient 0.8-billion-parameter model, reaches a Pearson correlation of 0.291 (Spearman 0.325) with human ratings on the seen styles and 0.381 on the unseen ones. On the seen styles it has the highest positive correlation among the metrics we test: Gram-matrix style loss and CLIP similarity correlate only weakly ($r=0.041$ and $r=0.168$), and LPIPS, scored so that a result closer to its source gets a higher score, is negatively correlated with human judgment ($r=-0.404$). We reach this result by distilling
evaluation knowledge from a much larger 72-billion-parameter vision-language judge, which we
use only during training to supervise \modelname{} and which itself reaches $r=0.327$,
confirming that most of the teacher's human-alignment survives compact distillation. To
train the system we build a dataset of source-to-stylized pairs across the same 12 artistic
styles using LoRA-adapted diffusion generation, and densify it with graded intermediate
examples produced by an image-morphing pipeline we adapt for higher-resolution generation,
so the evaluator learns to recognize partial as well as complete stylization. On two validation styles held out from training, \modelname{} reaches 28.2\% exact-match
agreement with its teacher's structured scores, against 1.1\% for the untuned base model,
while producing 67.8\% shorter reasoning traces, and on a single NVIDIA T4 it scores a pair in 39\,s (median) using 2.5\,GB of GPU memory. We release the human-rating benchmark, its annotation guideline, our SDXL-adapted FreeMorph implementation, and the model
weights so that future automatic metrics can be compared against the same human-correlation
criterion.
\end{abstract}




\maketitle

\section{Introduction}
\label{sec:intro}

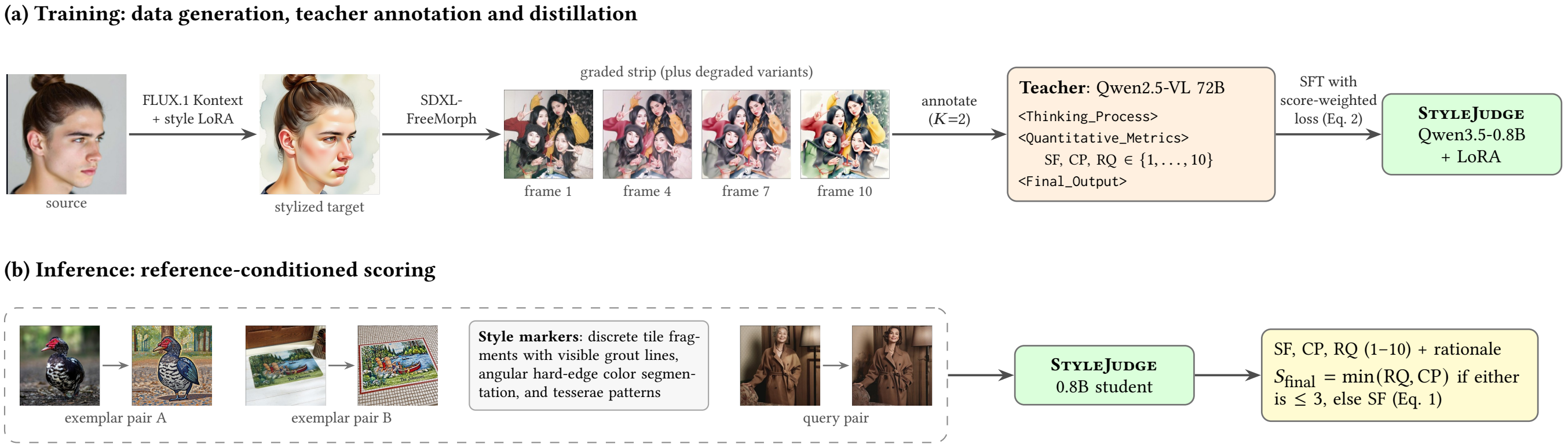
\begin{figure*}[t]
  \centering
  \resizebox{\textwidth}{!}{%
  \begin{tikzpicture}[
      font=\scriptsize,
      img/.style={inner sep=0pt, outer sep=1pt, draw=black!35, line width=0.3pt},
      card/.style={draw=black!55, rounded corners=3pt, fill=#1, inner sep=4pt, align=center, line width=0.5pt},
      icap/.style={font=\tiny, text=black!70, align=center, inner sep=1pt},
      arr/.style={-{Stealth[length=1.8mm]}, line width=0.7pt, black!70},
      sarr/.style={-{Stealth[length=1.2mm]}, line width=0.5pt, black!55},
      alab/.style={font=\tiny, text=black!85, align=center, above=1pt, inner sep=1pt},
      panel/.style={font=\footnotesize\bfseries, anchor=west, inner sep=0pt}
    ]
    \node[panel] at (0,2.35) {(a) Training: data generation, teacher annotation and distillation};
    \node[img, anchor=west] (src) at (0,1.0) {\includegraphics[width=1.35cm]{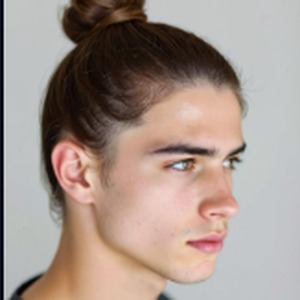}};
    \node[icap, below=0pt of src] {source};
    \node[img, right=1.45cm of src] (dst) {\includegraphics[width=1.35cm]{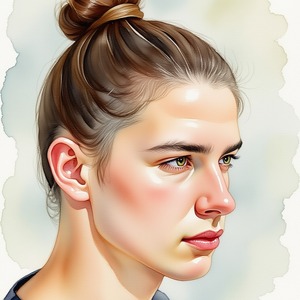}};
    \node[icap, below=0pt of dst] {stylized target};
    \draw[arr] (src.east) -- node[alab]{FLUX.1 Kontext\\+ style LoRA} (dst.west);
    \node[img, right=1.35cm of dst] (f1)  {\includegraphics[width=1.0cm]{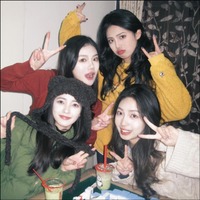}};
    \node[img, right=0.05cm of f1]  (f4)  {\includegraphics[width=1.0cm]{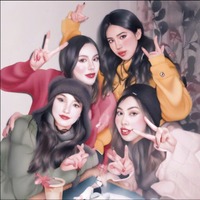}};
    \node[img, right=0.05cm of f4]  (f7)  {\includegraphics[width=1.0cm]{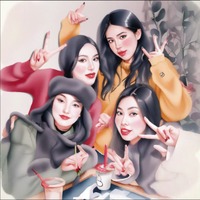}};
    \node[img, right=0.05cm of f7]  (f10) {\includegraphics[width=1.0cm]{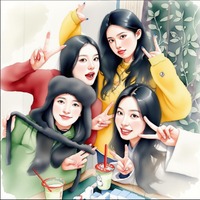}};
    \foreach \n/\k in {f1/1, f4/4, f7/7, f10/10} \node[icap, below=0pt of \n] {frame \k};
    \node[icap, above=1pt of $(f4.north)!0.5!(f7.north)$] {graded strip (plus degraded variants)};
    \draw[arr] (dst.east) -- node[alab]{SDXL-\\FreeMorph} (f1.west);
    \node[card=orange!12, right=1.3cm of f10, text width=2.75cm, align=left] (teach)
      {\textbf{Teacher}: Qwen2.5-VL 72B\\[2pt]
       {\tiny\ttfamily <Thinking\_Process>}\\
       {\tiny\ttfamily <Quantitative\_Metrics>}\\
       \hspace*{8pt}{\tiny SF, CP, RQ $\in\{1,\dots,10\}$}\\
       {\tiny\ttfamily <Final\_Output>}};
    \draw[arr] (f10.east) -- node[alab]{annotate\\($K{=}2$)} (teach.west);
    \node[card=green!14, right=1.2cm of teach, text width=1.75cm] (stu)
      {\textbf{\textsc{StyleJudge}}\\Qwen3.5-0.8B\\+ LoRA};
    \draw[arr] (teach.east) -- node[alab]{SFT with\\score-weighted\\loss (Eq.~\ref{eq:loss})} (stu.west);
    \node[panel] at (0,-0.55) {(b) Inference: reference-conditioned scoring};
    \node[img, anchor=west] (ea1) at (0.15,-1.62) {\includegraphics[width=0.9cm]{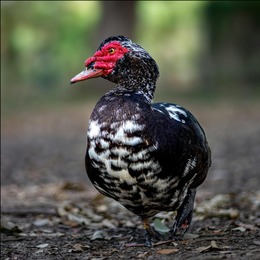}};
    \node[img, right=0.3cm of ea1] (ea2) {\includegraphics[width=0.9cm]{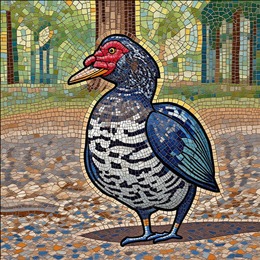}};
    \draw[sarr] (ea1.east) -- (ea2.west);
    \node[icap, below=0pt of $(ea1.south)!0.5!(ea2.south)$] (cea) {exemplar pair A};
    \node[img, right=0.32cm of ea2] (eb1) {\includegraphics[width=0.9cm]{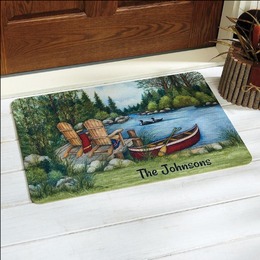}};
    \node[img, right=0.3cm of eb1] (eb2) {\includegraphics[width=0.9cm]{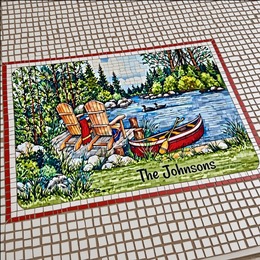}};
    \draw[sarr] (eb1.east) -- (eb2.west);
    \node[icap, below=0pt of $(eb1.south)!0.5!(eb2.south)$] (ceb) {exemplar pair B};
    \node[draw=black!35, rounded corners=2pt, fill=black!3, right=0.32cm of eb2, text width=2.5cm,
          align=left, inner sep=3pt, font=\tiny] (mk)
      {\textbf{Style markers}: discrete tile fragments with visible grout lines, angular hard-edge
       color segmentation, and tesserae patterns};
    \node[img, right=0.32cm of mk] (q1) {\includegraphics[width=0.9cm]{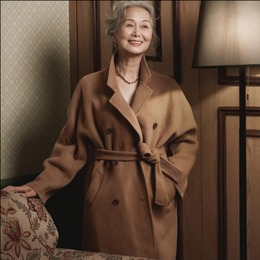}};
    \node[img, right=0.3cm of q1] (q2) {\includegraphics[width=0.9cm]{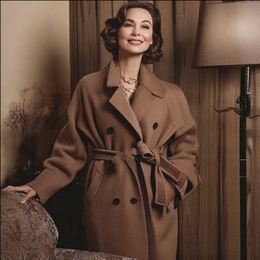}};
    \draw[sarr] (q1.east) -- (q2.west);
    \node[icap, below=0pt of $(q1.south)!0.5!(q2.south)$] (cq) {query pair};
    \begin{scope}[on background layer]
      \node[draw=black!45, dashed, rounded corners=4pt, inner sep=4pt,
            fit=(ea1)(ea2)(eb1)(eb2)(mk)(q1)(q2)(cea)(ceb)(cq)] (box) {};
    \end{scope}
    \node[card=green!14, right=0.75cm of box, text width=1.75cm] (mod)
      {\textbf{\textsc{StyleJudge}}\\0.8B student};
    \draw[arr] (box.east) -- (mod.west);
    \node[card=yellow!22, right=0.75cm of mod, text width=2.85cm, align=left] (out)
      {SF, CP, RQ (1--10) + rationale\\[2pt]
       $S_{\text{final}}=\min(\mathrm{RQ},\mathrm{CP})$ if either is $\le 3$, else SF (Eq.~\ref{eq:final_score})};
    \draw[arr] (mod.east) -- (out.west);
  \end{tikzpicture}}
  \caption{Overview of \modelname{}. (a)~Training: stylized targets from FLUX.1 Kontext and a style
  LoRA are expanded into graded ten-frame strips (plus degraded variants), annotated by a 72B teacher
  and distilled into a 0.8B student. (b)~Inference: two exemplar pairs, the text style markers and
  the query pair yield SF, CP and RQ, combined into $S_{\text{final}}$.}
  \Description{Two-row diagram. Top row, training: a source portrait is stylized into watercolor by FLUX.1 Kontext with a style LoRA; SDXL-FreeMorph expands it into a graded strip of frames; the Qwen2.5-VL 72B teacher annotates each item with XML reasoning and SF, CP, RQ scores; supervised fine-tuning with a score-weighted loss yields the StyleJudge student. Bottom row, inference: a prompt containing two mosaic exemplar pairs, the mosaic style markers and a query pair is scored by StyleJudge, which outputs SF, CP, RQ and a rationale combined into the final score.}
  \label{fig:pipeline}
\end{figure*}

Artistic style transfer is the task of transforming a content image into the visual idiom of a
reference artwork while preserving its semantic structure~\cite{gatys2016image}. Modern
diffusion-based generative models, combined with parameter-efficient Low-Rank Adaptation
(LoRA)~\cite{hu2022lora}, can now produce high-fidelity stylized outputs spanning
watercolor, cubism, charcoal, mosaic, and many other artistic domains. Despite this rapid
progress in generation quality, \textbf{automated evaluation of style transfer output
remains a fundamentally unresolved problem}.

A successful stylization must simultaneously satisfy two competing objectives: content
fidelity (\ie, preserving the identity, layout, and semantic structure of the input) and
style fidelity (\ie, faithfully expressing the distinguishing visual markers of the target
artistic domain). Human evaluation is the reference standard, but it is slow, expensive and hard to reproduce at scale. Existing automatic metrics each capture only a partial view of this trade-off:
Gram-matrix style loss~\cite{gatys2016image} captures global texture statistics but misses
compositional cues; LPIPS~\cite{zhang2018unreasonable} tends to penalise strong stylization
in favour of content retention; CLIP-based similarity~\cite{radford2021learning,hessel2021clipscore} does not distinguish
between weak and strong stylistic transformations; and DINO-based
features~\cite{caron2021emerging} are largely insensitive to brushstroke pattern, colour
shift, and abstraction level. We quantify how weakly, or in one case how negatively, the first three track human judgment on our benchmark (\S\ref{sec:main_results}); DINO features were not evaluated.

A common failure mode links all four: each metric operates in a \textbf{zero-shot manner},
evaluating an image pair without ever observing a reference example of the intended
transformation. Human evaluators, by contrast, implicitly calibrate to a reference before
rendering judgment. This observation is the starting point of our work: if the missing
ingredient is a reference example, an evaluator should be given one.

\paragraph{Proposed approach.} We introduce \modelname{}, an evaluation framework built
around \textbf{in-context learning}: rather than judging a query pair in isolation, the
evaluator is first shown a small number of high-quality reference source-to-stylized pairs,
turning evaluation into a visual analogy task the model can reason about directly. Running
such conditioning naively would require a large vision-language model at inference time,
which is too costly to run routinely; we therefore distill this in-context
reasoning from a large, accurate teacher into a compact 0.8-billion-parameter student that
retains most of the teacher's human alignment while running on a single 16\,GB GPU (NVIDIA T4). Figure~\ref{fig:pipeline} summarises the full pipeline.

\paragraph{Contributions.}
\begin{enumerate}
  \item[(i)] \modelname{}, a context-conditioned, teacher-distilled evaluator for artistic
        style transfer, together with the full pipeline used to train it: LoRA-based style
        generation, graded intermediate supervision via image morphing, and score-weighted
        distillation from a large teacher into a compact student (\S\ref{sec:method}),
        released together with our SDXL-adapted FreeMorph code; the 10K source images come from the public collections of Table~\ref{tab:base_dataset}.
  \item[(ii)] A direct validation of such a system against human ratings rather than only
        against its teacher: on a released, human-annotated benchmark of 878 rated pairs
        (728 on seen styles, 150 on two unseen styles), \modelname{} reaches $r{=}0.291$ on
        the seen and $r{=}0.381$ on the unseen styles, correlating better than Gram-matrix loss and CLIP similarity, while LPIPS correlates negatively; a per-style analysis shows that it ranks queries well within
        each style and that much of its remaining error is style-level calibration
        (\S\ref{sec:experiments}).
  \item[(iii)] A fully released, reproducible pipeline: the ratings with a manifest that maps
        every rating to its images, scoring scripts that reproduce the correlations of
        \S\ref{sec:main_results}, the training notebooks, our SDXL-adapted FreeMorph
        implementation, and the model weights~\cite{stylejudgecode2026}.
\end{enumerate}

\section{Related Work}
\label{sec:related}

\subsection{Style Transfer and Automated Quality Assessment}

Neural style transfer was introduced by Gatys \etal~\cite{gatys2016image} using Gram-matrix
feature matching in a VGG network. Subsequent work achieved real-time stylization via
feed-forward networks~\cite{johnson2016perceptual} and arbitrary style
conditioning~\cite{huang2017arbitrary}. Contemporary diffusion-based methods (\eg, FLUX.1-dev
Kontext~\cite{flux2025kontext}, IP-Adapter~\cite{ye2023ip}) paired with LoRA fine-tuning now
produce high-fidelity style transfer across diverse artistic domains, which is precisely
what makes reliable evaluation of their output urgent. Yet the metrics available were
largely not designed for it: general-purpose image quality assessment (IQA) metrics such as
SSIM~\cite{wang2004image}, LPIPS~\cite{zhang2018unreasonable}, and FID~\cite{heusel2017gans}
target photorealism or distortion, not stylistic transformation, while style-specific losses
(\eg, Gram matrices, style CLIP) and CLIP-IQA~\cite{wang2023exploring} capture only
surface-level statistics or conflate content alignment with stylistic faithfulness.
Style-specific measures have also been proposed: ArtFID~\cite{wright2022artfid} combines a
content distance with a set-level Fr\'echet distance, and CSD~\cite{somepalli2024csd} learns
contrastive style descriptors for image-to-image style similarity. None of these conditions
on an example of the intended transformation.

\subsection{Vision-Language Models as Evaluators}

Recent work has explored vision-language models (VLMs) as flexible evaluators for image
generation tasks~\cite{ku2023viescore,wu2023human}. ImgEdit-Judge~\cite{pkuimgedit} fine-tunes Qwen2.5-VL for image
editing assessment, and LLM-as-a-judge frameworks~\cite{zheng2023judging} have likewise been
adapted to the multimodal domain and surveyed in the text-only setting~\cite{gu2024surveyllmasajudge}. Our work
differs by operating in an in-context, rather than zero-shot, regime; by targeting the
specific domain of artistic style transfer; by distilling a large judge into a compact
deployable model; and by validating the result directly against human judgment rather than
only against the teacher model. Human-preference models such as HPSv2~\cite{wu2023human}, ImageReward~\cite{xu2023imagereward} and PickScore~\cite{kirstain2023pickapic} are trained on human preferences for text-to-image generation, but they do not condition on a reference style transformation.

\subsection{In-Context Learning and Knowledge Distillation}

In-context learning (ICL) was shown to dramatically improve large language model performance
on diverse tasks by conditioning on a small number of labelled
examples~\cite{brown2020language}; Flamingo~\cite{alayrac2022flamingo} demonstrated the analogous few-shot capability in the multimodal setting, and LLaVA~\cite{liu2023visual} and PandaGPT~\cite{su2023pandagpt} extended vision-language models to general instruction following. More recent empirical work has questioned how much of ICL's apparent few-shot
generalisation reflects genuine learning from the provided exemplars versus reliance on a
model's prior knowledge, finding that its ability to generalise to unseen tasks is often
limited~\cite{dewynter2025icllearning}. This motivates the controlled ablation over the
number of exemplar pairs we run in \S\ref{sec:ablation}, rather than assuming that more
in-context reference pairs necessarily yield better calibration. To our knowledge, ICL has
not previously been applied specifically to artistic style transfer evaluation, nor
validated against human ratings on a dedicated benchmark in this domain. Separately,
distilling from large teacher models to smaller students via supervised fine-tuning on
teacher-generated rationales has proven effective across many
domains~\cite{ho2022large,west2022symbolic}; we apply this paradigm to multimodal style
assessment, both to retain the teacher's alignment with human judgment and to avoid the GPU
cost and latency of running a 72B judge at inference time.

\section{Proposed Method}
\label{sec:method}

\subsection{Reference-Conditioned Evaluation Protocol}
\label{sec:eval_protocol}

Each \modelname{} evaluation instance consists of (1) a small number of in-context reference
pairs, each a high-quality source-to-stylized example of the target style treated as a
``perfect'' exemplar, and (2) a query pair: a new source image and a candidate stylized
output to be judged. The reference pairs calibrate the model's style-specific scoring prior
before it evaluates the query, converting evaluation into a visual analogy task. We adopt two
reference pairs throughout ($K=2$), a choice we examine in \S\ref{sec:ablation}. Each prompt also carries a short text description of the target style (three visual markers, written with web search and an LLM), used in training and at test time (Table~\ref{tab:lora_adapters}), so the conditioning is not exemplar-only.

Given a query pair, the evaluator produces three independent 1--10 integer scores rather
than a single style-fidelity number, so that a failure of rendering or content preservation
cannot masquerade as a stylistic success: Style Fidelity (SF), the degree to which
target-style visual markers are present globally throughout the image; Content Preservation
(CP), the integrity of the original scene's objects, layout, and fine structural detail; and
Rendering Quality (RQ), the technical fidelity of the output, \ie, the absence of artefacts
and the presence of coherent textures and clean transitions. These three scores are combined
into a single composite score,

\begin{equation}
  S_{\text{final}} =
  \begin{cases}
    \min(\mathrm{RQ},\,\mathrm{CP}) & \text{if } \mathrm{RQ}\!\le\!3
                                       \text{ or } \mathrm{CP}\!\le\!3, \\[4pt]
    \mathrm{SF}  & \text{otherwise.}
  \end{cases}
  \label{eq:final_score}
\end{equation}

The rationale is discriminability, not weighting. When RQ or CP falls to $\le\!3$, rendering
artefacts or content destruction are severe enough that the query pair can no longer be
visually compared against the reference transformation at all: the evaluator cannot tell
\emph{how} the style changed because the underlying image itself is compromised, so
$S_{\text{final}}$ is clamped to the more severe of the two failures and Style Fidelity is
not a meaningful signal. Otherwise the transformation is legible and $S_{\text{final}}$ reduces to SF, the quantity the evaluation is meant to measure.
This threshold-based design reflects a pattern we observed while piloting the protocol with
human annotators (\S\ref{sec:pilot}): raters treated rendering and content failures as
blocking conditions rather than as factors to be traded off against style quality.

To make these scores reliably extractable, and to reduce the framing bias a free-text prompt can introduce in an LLM-as-judge setting, the evaluator is constrained to a three-block XML output: \texttt{<Thinking\_Process>} (style extraction from the exemplars, an analogy check against the query, and a conflict check for artefacts), \texttt{<Quantitative\_Metrics>} (SF, CP, RQ), and \texttt{<Final\_Output>} (a brief rationale).

\subsection{Training Data Generation}
\label{sec:data_gen}

Training \modelname{} requires a large set of source-to-stylized pairs with graded, not just
binary, examples of stylization intensity. We generate stylized targets using the
\textbf{FLUX.1-dev Kontext} diffusion model~\cite{flux2025kontext} (24 sampling steps) together with
style-specific LoRA adapters sourced from the fal.ai community
collection~\cite{falkontext2025}, covering twelve visually distinct artistic styles
(Table~\ref{tab:lora_adapters}).
LoRA~\cite{hu2022lora} adapters allow lightweight style specialisation and rapid adapter swapping during dataset construction.

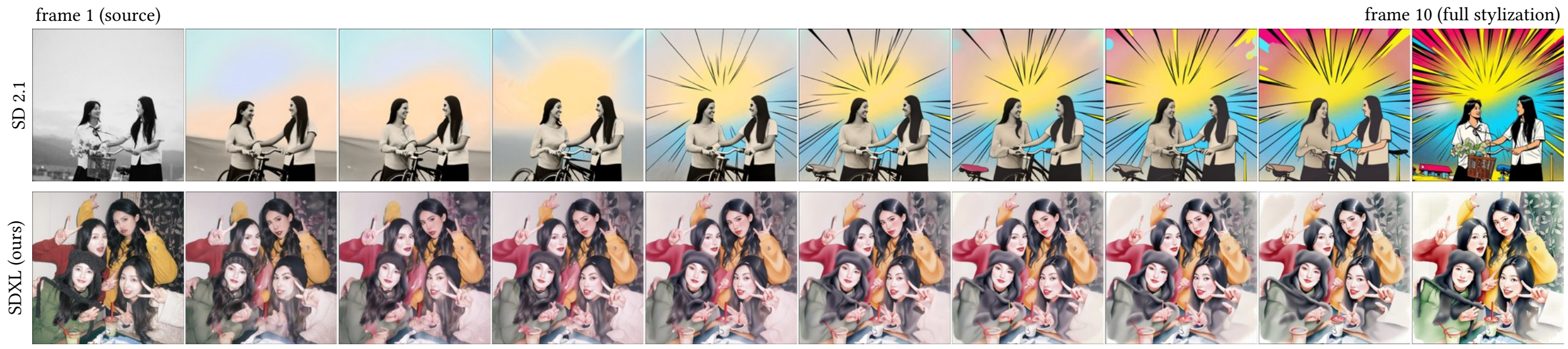
\begin{figure*}[t]
  \centering
  \begin{tikzpicture}[font=\scriptsize]
    \node[inner sep=0pt] (r1) {\includegraphics[width=0.955\textwidth]{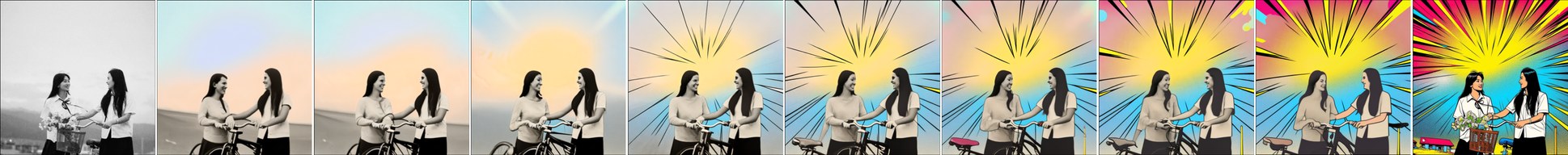}};
    \node[inner sep=0pt, below=3pt of r1] (r2) {\includegraphics[width=0.955\textwidth]{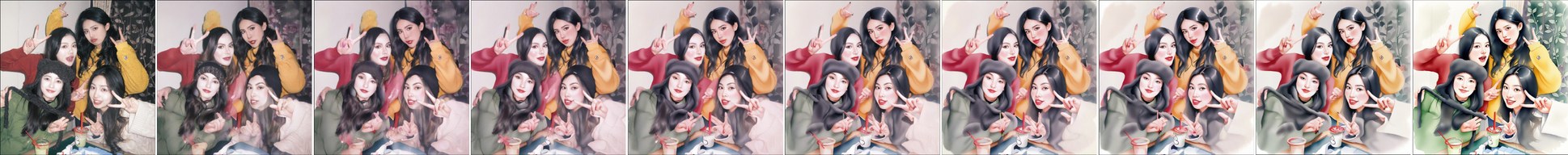}};
    \node[rotate=90, anchor=south, inner sep=2pt] at (r1.west) {SD 2.1};
    \node[rotate=90, anchor=south, inner sep=2pt] at (r2.west) {SDXL (ours)};
    \node[anchor=south west, inner sep=1pt] at (r1.north west) {frame 1 (source)};
    \node[anchor=south east, inner sep=1pt] at (r1.north east) {frame 10 (full stylization)};
  \end{tikzpicture}
  \caption{Morph strips from the original SD~2.1 FreeMorph (top) and our SDXL adaptation (bottom),
  frames 1 to 10 (different input pairs). SD~2.1 changes objects and clothing alongside the style;
  SDXL changes the rendering while keeping identity and layout stable. More strips are in the
  project repository~\cite{stylejudgecode2026}.}
  \Description{Two rows of ten frames each. The top row, from SD 2.1 FreeMorph, turns a black-and-white photo of two women with a bicycle into a pop-art image, with objects and clothing changing between frames. The bottom row, from the SDXL adaptation, gradually turns a group photo into a painterly rendering while the people and layout stay the same.}
  \label{fig:morph}
\end{figure*}

A model trained only on binary source-to-fully-stylized pairs would learn a coarse,
all-or-nothing notion of style intensity. We therefore add \textbf{densely graded
intermediate supervision}: for every source-target pair we generate eight interpolated frames using a morphing pipeline adapted from FreeMorph~\cite{freemorphcite}, giving a ten-frame strip from the source (frame~1) to the full stylization (frame~10), so that the model can (i) distinguish subtle stylization intensity (frames 1--4), (ii)
identify when stylization is deep enough to trade off minor content fidelity (frames 5--7),
and (iii) recognise a fully faithful transfer (frames 8--10). FreeMorph's native Stable
Diffusion 2.1 backbone introduced undesirable object-level warping rather than
style-progressive evolution, changing object shapes and composition alongside style, which
is unsuitable for measuring style intensity independently of content drift; we therefore
re-target the pipeline to SDXL~\cite{podell2023sdxl}, with dual text-encoder conditioning
(interpolating pooled CLIP-G and CLIP-L embeddings across frames for semantic consistency)
and reduced edit strength and classifier-free guidance scale that together prioritise
semantic consistency over structural transformation. Relative to the SD~2.1 setting (768\,px, edit strength 0.8, guidance 7.5), we use 1024\,px, strength 0.6 and guidance 6.5, with 50 denoising steps; latents are interpolated spherically and prompt and pooled text embeddings linearly. Figure~\ref{fig:morph} contrasts the two pipelines.

\textbf{Failure-mode supervision.} Graded strips rarely contain the severe rendering or content
failures that the clamp of Eq.~\ref{eq:final_score} is designed to catch. We therefore also
generate deliberately degraded sequences with a corrupted variant of the morphing pipeline:
the guidance scale is raised from 6.5 to 25, the negative prompt is replaced with one that
penalises quality, the frequency-filter thresholds are set close to zero, and Gaussian noise
is added to the intermediate latents. These sequences expose the evaluator, and the human
raters (\S\ref{sec:setup}), to outputs that fail on rendering or content.

\begin{table*}[t]
\caption{The twelve styles and the text style markers included in every prompt (written with web
search and an LLM; also given to the human raters as style definitions). Acrylic and Pop Art
($^{\dagger}$) are held out from training and used only as unseen test styles (\S\ref{sec:setup}).}
\label{tab:lora_adapters}
\scriptsize
\begin{tabularx}{\textwidth}{@{}l X l X@{}}
\toprule
\textbf{Style} & \textbf{Style markers} & \textbf{Style} & \textbf{Style markers} \\
\midrule
Abstract & non-representational geometric forms, bold flat primary color blocks, and elimination of figurative elements & Impressionist & short broken dappled brushstrokes, chromatic color juxtaposition without blending, and soft atmospheric hazy edges \\
Acrylic$^{\dagger}$ & thick impasto surface texture, sharp-edged opaque brushwork, and high-saturation vivid colors & Minimalist & extreme negative white-space areas, a single dominant restrained color, and reductive simplified forms \\
Charcoal & monochromatic grey tonal gradients, smudged dusty texture marks, and visible hatching lines & Mosaic & discrete tile fragments with visible grout lines, angular hard-edge color segmentation, and tesserae patterns \\
Cubist & simultaneous multiple-viewpoint planes, fractured interlocking geometric facets, and a muted palette & Pencil & parallel graphite hatching lines, linear tonal gradation from light to dark, and visible paper grain texture \\
Expressive & raw emotionally distorted brushwork, unblended aggressive color strokes, and angular forms & Pop Art$^{\dagger}$ & flat bold primary poster colors, Ben-Day halftone dot patterns, and thick black outline contour lines \\
Gouache & opaque chalky flat color areas, matte non-reflective finish, and clean sharp color edges & Watercolor & transparent layered wash glazes, soft wet-edge bloom blending, and paper granulation in wash areas \\
\bottomrule
\end{tabularx}
\end{table*}

\subsection{Teacher Annotation and Score-Weighted Distillation}
\label{sec:distillation}

Every generated pair, including intermediate frames, is annotated by a large teacher model,
Qwen2.5-VL 72B~\cite{bai2025qwen25vl}, chosen for its strong published benchmarks on visual
question answering, multimodal reasoning, and instruction following, and for being openly
available at this scale, unlike closed frontier VLMs, which lets us run it locally at the
volume our dataset requires without per-call API cost. It produces the full three-block XML
output described in \S\ref{sec:eval_protocol} under the same protocol the student will later
use. These teacher rationales become the supervised fine-tuning targets for a compact
student, Qwen3.5-0.8B~\cite{qwenteam2025qwen35}, fine-tuned with LoRA adapters; Qwen2.5-VL serves only as the teacher.

Standard cross-entropy distributes equal loss weight across all output tokens. Given that the
critical scoring tokens, the digits inside \texttt{<Quantitative\_Metrics>}, are only a few tokens of an output that runs to hundreds of tokens, we apply a positional up-weighting scheme,
\begin{equation}
  \mathcal{L} = \frac{\sum_{t \in \mathcal{T}_{\text{act}}} w_t \cdot \ell_t}
                     {|\mathcal{T}_{\text{act}}|}, \quad
  w_t =
  \begin{cases}
    1 + \lambda & \text{score digit in last 30\%,} \\
    1           & \text{otherwise,}
  \end{cases}
  \label{eq:loss}
\end{equation}
where $\lambda$ is a score weight and $\mathcal{T}_{\text{act}}$ excludes all padding and
image tokens; \S\ref{sec:ablation} reports the sweep over $\lambda$ that fixes its value at $5.0$. 

\textbf{Implementation.} The teacher (AWQ-quantised) is served with vLLM. The student
(853M parameters, about 1.7\,GB of weights in bfloat16) is fine-tuned with Unsloth for one
epoch: LoRA rank 16
and $\alpha{=}32$ on all attention and MLP projections of both the vision and the language
layers, learning rate $2{\times}10^{-4}$ with cosine decay and 10 warm-up steps, an effective
batch size of 16, bfloat16 precision and an 8{,}192-token context; images are loaded at 1024 pixels and resized to 512 pixels by Unsloth's vision pipeline. Complete
settings and the training notebooks are in the project repository~\cite{stylejudgecode2026}.

\section{Experiments}
\label{sec:experiments}

\subsection{Zero-Shot Pilot Study}
\label{sec:pilot}

Before building around in-context conditioning, we ran a small pilot to confirm that
zero-shot VLM evaluation genuinely leaves a gap to human judgment. We used
Qwen2.5-VL~\cite{bai2025qwen25vl} 7B fine-tuned with ImgEdit-Judge~\cite{pkuimgedit} weights,
scoring each of 88 source-stylized pairs from ImgEdit (whose styles differ from our twelve) on the same three dimensions used throughout this
paper. Of these, 50 had valid paired annotations from two raters, both project-team members (one of whom also rated the main human benchmark, \S\ref{sec:setup}) and were used for correlation analysis. The pilot's human-human inter-rater agreement ($r=0.25$, 95\% CI $[-0.03, 0.49]$, $p=0.08$) cannot be distinguished from zero at $n=50$, so we report it only as an indicative reference, not as a ceiling. Table~\ref{tab:zeroshot} summarises the result: the model correlates only weakly with human ratings ($r=0.19$) and systematically under-estimates scores in most cases.

\begin{table}[t]
\caption{Quantitative evaluation metrics: Zero-Shot VLM vs.\ Human annotators.}
\label{tab:zeroshot}
\begin{tabular}{@{}lp{3.6cm}@{}}
\toprule
\textbf{Metric} & \textbf{Result} \\
\midrule

Mean score (VLM / human) & $5.86 \pm 1.66$ / $6.62 \pm 1.95$ \\
Pearson correlation ($r$) & $0.19$ \quad ($p=0.18$) \\
Spearman correlation ($\rho$) & $0.11$ \quad ($p=0.46$) \\
Human inter-rater ($r$) & $0.25$ (95\% CI $[-0.03,0.49]$, $p{=}0.08$) \\
VLM systematic bias & Under-estimated in 62\% of cases (31/50) \\

\bottomrule
\end{tabular}
\end{table}

Qualitatively, without reference examples the zero-shot model collapses toward generic image quality assessment: it penalises successful abstraction and agrees best on visually explicit styles, the gap the in-context design of \S\ref{sec:method} is built to close.

\subsection{Dataset, Splits, and Human Evaluation Protocol}
\label{sec:setup}

\begin{figure*}[t]
  \centering
  \setlength{\tabcolsep}{0.6pt}\renewcommand{\arraystretch}{0.6}
  \begin{tabular}{@{}cccccccccccc@{}}
\includegraphics[width=0.0795\textwidth]{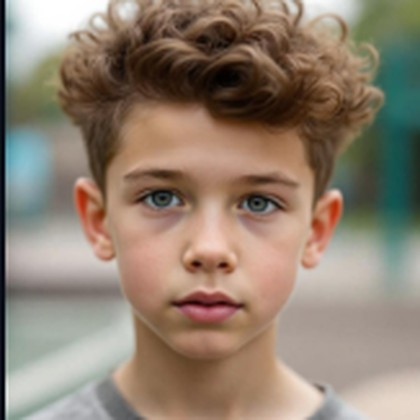} & \includegraphics[width=0.0795\textwidth]{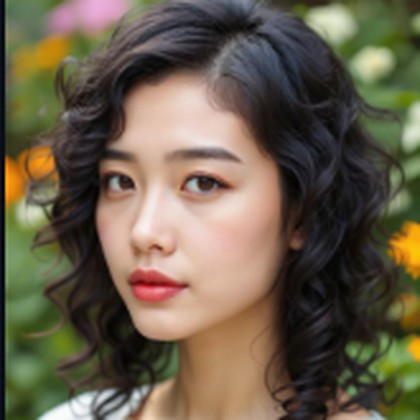} & \includegraphics[width=0.0795\textwidth]{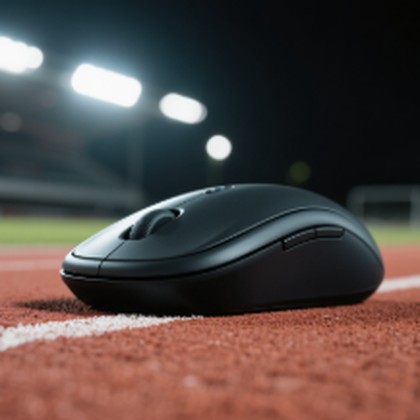} & \includegraphics[width=0.0795\textwidth]{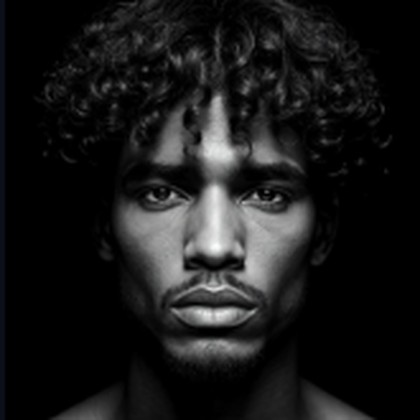} & \includegraphics[width=0.0795\textwidth]{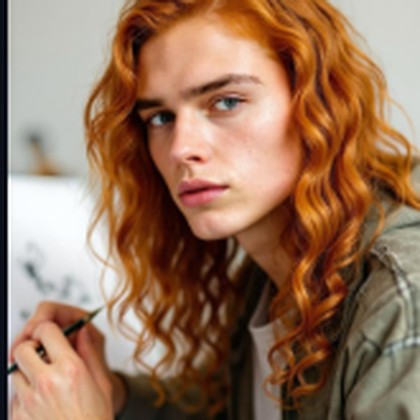} & \includegraphics[width=0.0795\textwidth]{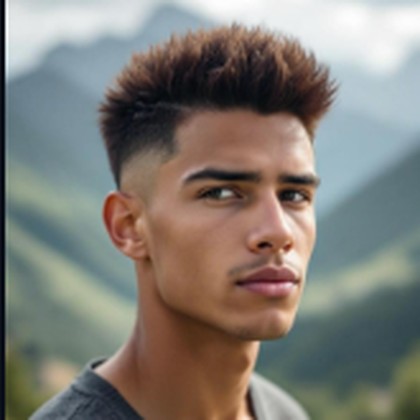} & \includegraphics[width=0.0795\textwidth]{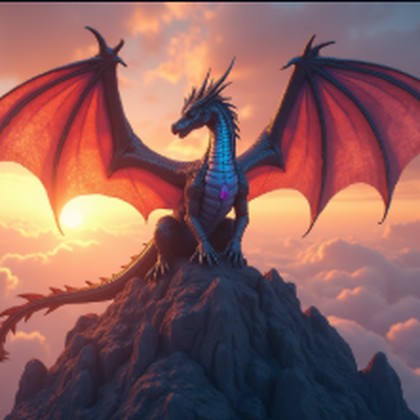} & \includegraphics[width=0.0795\textwidth]{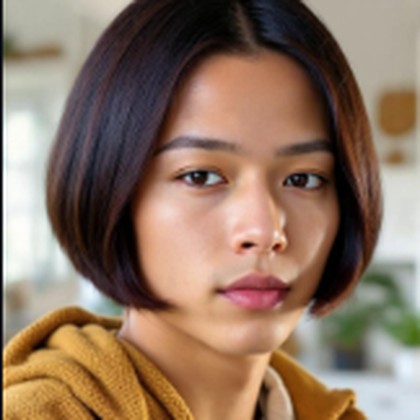} & \includegraphics[width=0.0795\textwidth]{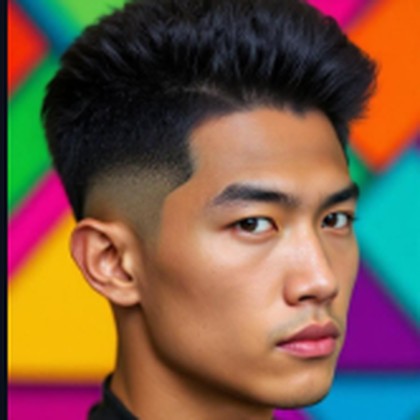} & \includegraphics[width=0.0795\textwidth]{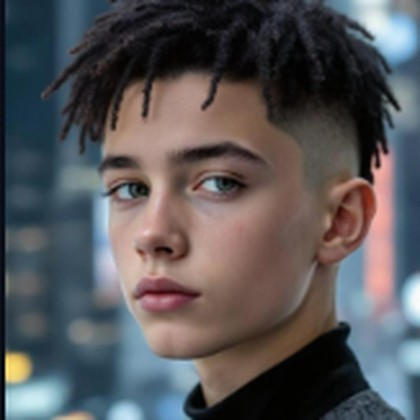} & \includegraphics[width=0.0795\textwidth]{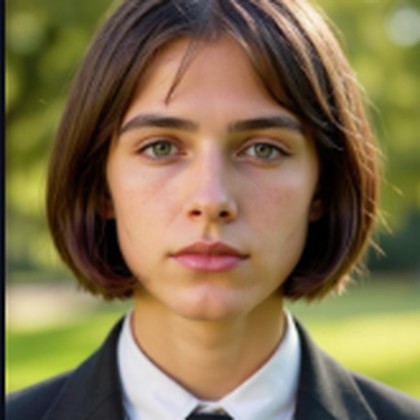} & \includegraphics[width=0.0795\textwidth]{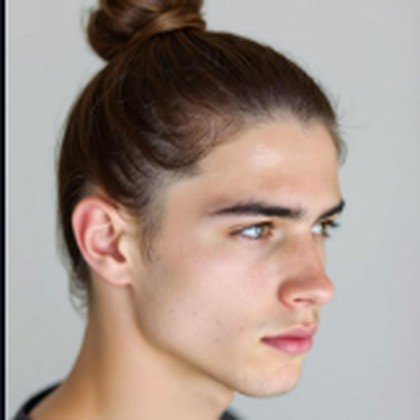} \\
\includegraphics[width=0.0795\textwidth]{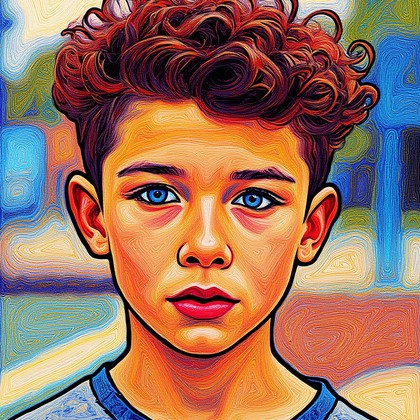} & \includegraphics[width=0.0795\textwidth]{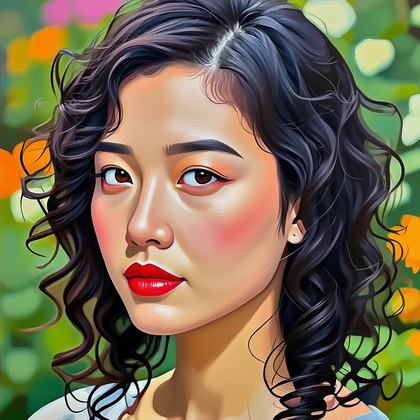} & \includegraphics[width=0.0795\textwidth]{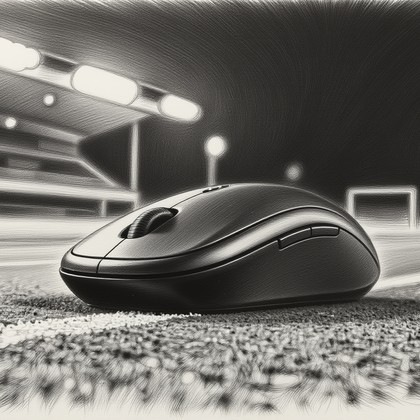} & \includegraphics[width=0.0795\textwidth]{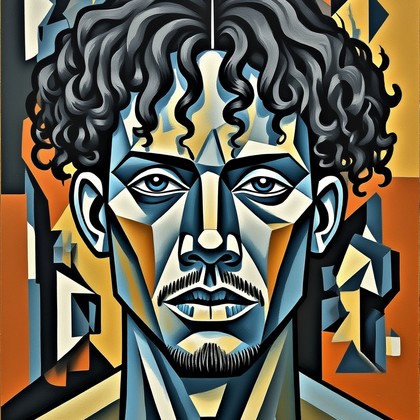} & \includegraphics[width=0.0795\textwidth]{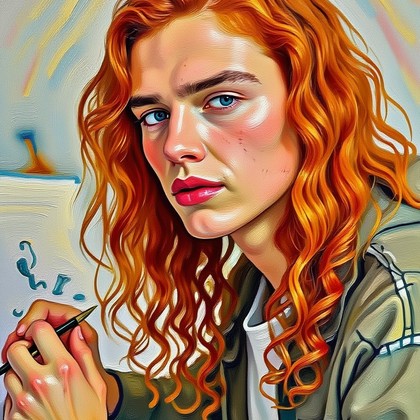} & \includegraphics[width=0.0795\textwidth]{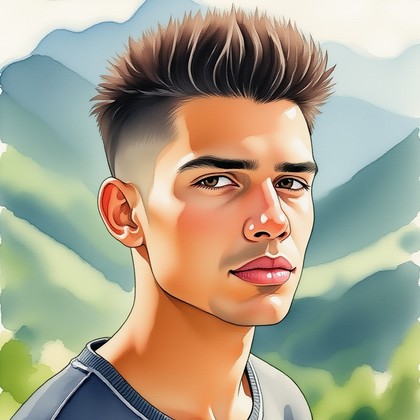} & \includegraphics[width=0.0795\textwidth]{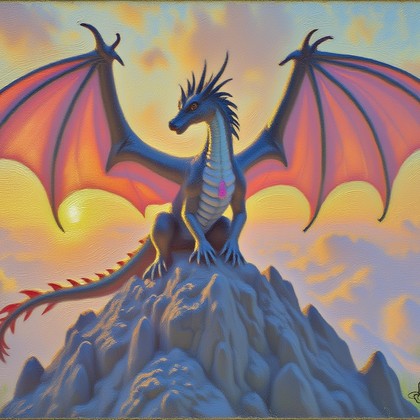} & \includegraphics[width=0.0795\textwidth]{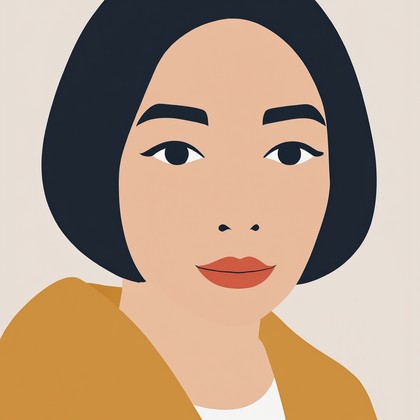} & \includegraphics[width=0.0795\textwidth]{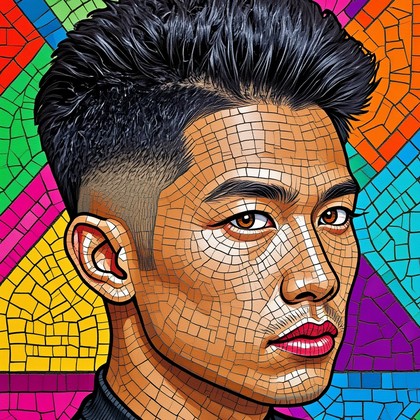} & \includegraphics[width=0.0795\textwidth]{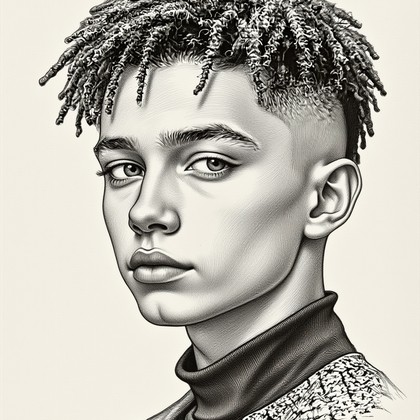} & \includegraphics[width=0.0795\textwidth]{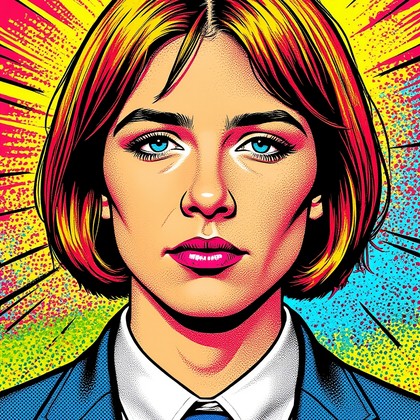} & \includegraphics[width=0.0795\textwidth]{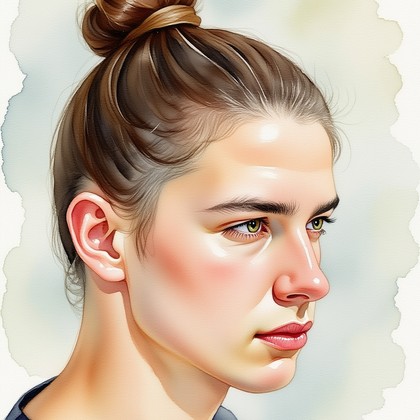} \\[2pt]
{\scriptsize Abstract} & {\scriptsize Acrylic$^{\dagger}$} & {\scriptsize Charcoal} & {\scriptsize Cubist} & {\scriptsize Expressive} & {\scriptsize Gouache} & {\scriptsize Impressionist} & {\scriptsize Minimalist} & {\scriptsize Mosaic} & {\scriptsize Pencil} & {\scriptsize Pop Art$^{\dagger}$} & {\scriptsize Watercolor} \\
  \end{tabular}
  \caption{Source images (top) and FLUX.1-dev Kontext stylized outputs (bottom) for each of the twelve
  style LoRAs. $^{\dagger}$Unseen test styles.}
  \Description{A row of twelve image pairs, one per style, each showing a source photograph above its stylized version.}
  \label{fig:style_grid}
\end{figure*}

Source images were aggregated from seven complementary collections spanning general,
non-photorealistic, macro/natural-history, and landscape content (Table~\ref{tab:base_dataset}).
Only square images were retained and all were standardised to $1024{\times}1024$ pixels via
Lanczos resampling: for HQ-50K, images with side lengths in $[800,1024)$ pixels were
upscaled, and for all other sources the threshold was $[1000,1024)$; images exceeding
1{,}024 pixels on either side were downscaled, and images below the respective lower
threshold were discarded. This yielded a curated pool of approximately 10K images, stylized
across the twelve LoRA adapters of \S\ref{sec:data_gen} to produce source-stylized pairs,
each further densified by the eight intermediate frames described there.
Figure~\ref{fig:style_grid} shows representative pairs for every style.

\begin{table}[t]
\caption{Source collections used to build the curated 10K-image pool.}
\label{tab:base_dataset}
\small
\begin{tabular}{@{}>{\raggedright\arraybackslash}p{2.3cm}p{5.1cm}@{}}
\toprule
\textbf{Source} & \textbf{Contribution} \\
\midrule
HQ-50K~\cite{hq50k} & Indoor, outdoor, artificial-domain coverage \\
LAION~\cite{schuhmann2022laion,laion1024} & Large-scale, general $1024{\times}1024$ pool \\
Omni\-Consistency~\cite{omniconsistency} & Non-photorealistic: 3D, Chibi, illustration \\
Pexels~\cite{pexels}, iNaturalist~\cite{vanhorn2018inaturalist,inat2021birds,inatfish,inatinsects} & Macro subjects: insects, birds, fish, botanicals \\
LHQ-1024~\cite{lhq1024,skorokhodov2021aligning}, Unsplash~\cite{unsplash} & Landscape scenes, atmospheric variation \\
\bottomrule
\end{tabular}
\end{table}

Table~\ref{tab:data_split} summarises the resulting split for teacher-rationale training:
eight styles are used for training, two held-out (in-distribution) styles for validation, and
two entirely unseen styles for testing (Acrylic and Pop Art, marked in
Table~\ref{tab:lora_adapters}). Teacher agreement (\S\ref{sec:teacher_accuracy}) is measured on the validation styles (expressive, mosaic), and the 150 unseen-style human ratings (\S\ref{sec:main_results}) come from the test styles; neither is used during fine-tuning (the main 728-pair human benchmark uses the ten training and validation styles).

\begin{table}[t]
\caption{Splits of the teacher-annotated data used to train the $K{=}2$ student.}
\label{tab:data_split}
\begin{tabular}{@{}lrrp{2.8cm}@{}}
\toprule
\textbf{Split} & \textbf{Samples} & \textbf{Styles} & \textbf{Note} \\
\midrule
Train & 5{,}614 & 8 & Teacher rationales \\
Val   & 1{,}326 & 2 & Expressive, mosaic \\
Test  & 800 & 2 & \textit{Unseen}: Acrylic, Pop Art \\
\bottomrule
\end{tabular}
\end{table}

Separately from these teacher-rationale splits, we collected 878 human ratings (four empty rows were discarded) over source-stylized query pairs from three raters, members of the project team with no formal art training, who rated independently of one another; before rating, they were briefed with the style definitions of Table~\ref{tab:lora_adapters} and example images of each style, and each pair was rated by one rater (502, 276 and 100 pairs). During annotation the interface showed the two exemplar pairs, the query pair, a 1--10 score selector and a free-text field for a short justification. The guideline, shown verbatim to every rater before annotation, mirrors Eq.~\ref{eq:final_score}: \textit{``If rendering or content preservation is very poor (for example, you cannot discriminate the style transfer or the output is severely degraded), give a score between 1--3. If rendering and content preservation are almost good, then decide the score solely on the basis of style fidelity (\ie, how well the output matches the target style), using the most intuitive judgement possible. Provide a short reasoning statement explaining the logic behind your score.''} Of the 878 pairs, 728 come from the ten training and validation styles and 150 from the two unseen styles (75 each); we call the ten \emph{seen} (the two validation styles are held out from training but were used for model selection) and Acrylic and Pop Art \emph{unseen}.  Query pairs are disjoint from the training pairs. To cover the quality range evenly, items were sampled with approximately uniform teacher scores: 100 items per score level from 1 to 9, except 78 at level~6, because 22 sampled items at that level were not rated. Query results come from the graded morph strips (643 pairs) and from three further generation batches, including deliberately degraded outputs (235 pairs; the source of each query is listed in the released manifest), and the human ratings use the full 1--10 scale (mean 5.4). This gives $N{=}728$ matched (human rating, model prediction) pairs for the seen-style analysis in \S\ref{sec:main_results}. The full rating CSV and guideline are released with the benchmark. 

\subsection{Main Results: Correlation with Human Judgment}
\label{sec:main_results}

Table~\ref{tab:human_correlation} reports Pearson and Spearman correlation with this
human-validated subset for the three traditional automatic metrics introduced in
\S\ref{sec:related} and for \modelname{} at both the teacher and distilled-student stage.
Gram-matrix style loss~\cite{gatys2016image} is the mean squared difference between the Gram matrices, \ie, the channel-wise feature correlations, of VGG features of the candidate output and the exemplar's stylized image; LPIPS (VGG) compares the query source with its result; and CLIP similarity is the cosine similarity of CLIP ViT-B/32 embeddings of the result and the exemplar's stylized image. Both distances are negated, so that a higher score means a closer match, and all three metrics use $256{\times}256$ images and are min--max scaled to 1--10 and rounded. Gram loss is statistically indistinguishable from zero correlation and CLIP similarity captures only weak semantic alignment, while LPIPS is negatively correlated ($r=-0.404$; $+0.405$ for the unrounded raw distance), with a larger magnitude than \modelname{}'s positive correlation. This is expected of a content-preservation metric: LPIPS distance to the source grows with stylization strength along our morph strips (Spearman $\rho=0.66$ with frame index), whereas the raters reward stronger stylization. The distilled 0.8B \modelname{} student reaches $r=0.291$ (95\% bootstrap CI $[0.225, 0.351]$) against $0.327$ ($[0.258, 0.389]$) for its 72B teacher; the difference is not significant (CI $[-0.112, 0.037]$). The untuned 0.8B base model is uncorrelated with human ratings zero-shot ($r=0.019$ on the 711 pairs it scores), so the gain comes from fine-tuning. On the 150 unseen-style pairs, where exemplars and style markers are still supplied, the student reaches $r=0.381$ (Spearman $0.436$; bootstrap CI $[0.24, 0.51]$) and the teacher $0.408$; over all 878 pairs they reach $0.297$ and $0.341$. Per-rater student correlations are positive for all three raters but modest ($0.333$, $0.124$, $0.109$ on 419, 232 and 77 pairs). Each pair has a single rater, so no same-set inter-rater ceiling is available and we make no claim of human parity.  All correlations and intervals in this section, including the per-style analysis below, are reproduced by the scripts released with the benchmark (bootstrap with 2{,}000 resamples).

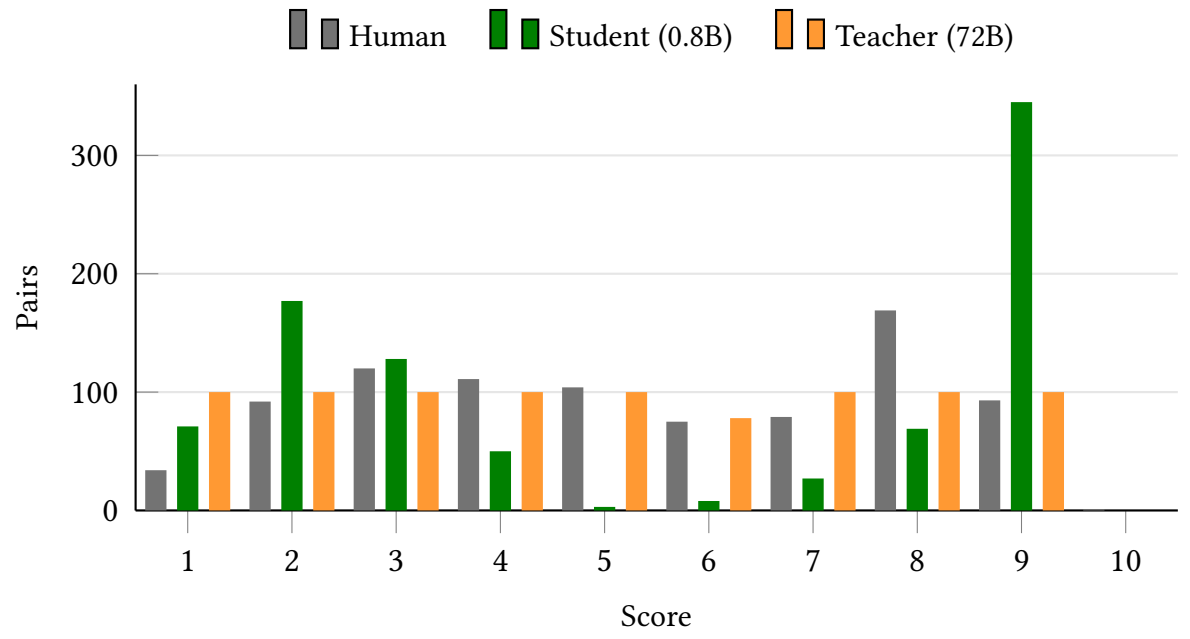
\begin{figure}[t]
  \centering
  \begin{tikzpicture}
    \begin{axis}[
        width=\columnwidth, height=4.4cm,
        ybar, bar width=4pt,
        xmin=0.5, xmax=10.5, xtick={1,...,10},
        ymin=0, ymax=360, ytick={0,100,200,300},
        xlabel={Score}, ylabel={Pairs},
        label style={font=\scriptsize}, tick label style={font=\scriptsize},
        legend style={font=\scriptsize, draw=none, at={(0.5,1.02)}, anchor=south, legend columns=3,
                      /tikz/every even column/.append style={column sep=6pt}},
        ymajorgrids, grid style={black!10}, axis lines*=left]
      \addplot[fill=black!55, draw=none] coordinates
        {(1,34) (2,92) (3,120) (4,111) (5,104) (6,75) (7,79) (8,169) (9,93) (10,1)};
      \addlegendentry{Human}
      \addplot[fill=green!50!black, draw=none] coordinates
        {(1,71) (2,177) (3,128) (4,50) (5,3) (6,8) (7,27) (8,69) (9,345) (10,0)};
      \addlegendentry{Student (0.8B)}
      \addplot[fill=orange!80, draw=none] coordinates
        {(1,100) (2,100) (3,100) (4,100) (5,100) (6,78) (7,100) (8,100) (9,100) (10,0)};
      \addlegendentry{Teacher (72B)}
    \end{axis}
  \end{tikzpicture}
  \caption{Score distributions on the 878 benchmark pairs. Human ratings spread across the
  scale; the student's scores pile up at 9 (39\%) and at 1--3 (43\%), with few in the middle;
  the teacher's are flat by construction, because items were sampled uniformly over its score.}
  \Description{Grouped bar chart of how often each score from 1 to 10 occurs for humans, the student and the teacher. Human ratings are spread out with a peak at 8; the student's scores are concentrated at 9 and at 1 to 3; the teacher's are about 100 per score level from 1 to 9.}
  \label{fig:score_dist}
\end{figure}

\textbf{Per-style analysis.} Table~\ref{tab:per_style} breaks the correlations down by style.
Both models have positive point estimates in all twelve styles (for mosaic, and for the teacher on minimalist, the 95\% intervals include zero); the student's per-style $r$ ranges from $0.14$ (mosaic) to $0.57$ (cubist) and exceeds the
teacher's in 10 of the 12 styles. Within a style, the student ranks queries better than the
pooled number suggests: after subtracting each style's mean from both the human and the model
scores, its correlation on the 728 seen-style pairs rises to $r=0.368$ (teacher $0.333$). What
lowers the pooled correlation is style-level calibration: the student's mean score ranges from
2.7 (Pop Art) to 7.8 (watercolor), while the human means stay between 4.8 and 6.4. Figure~\ref{fig:score_dist} shows the same effect in the score distributions: the student uses the middle of the scale (4--6) for only 7\% of pairs, whereas 33\% of human ratings fall there. A per-style calibration of the student's scores is therefore a cheap and promising improvement.

\begin{table}[t]
\caption{Per-style Pearson $r$ with human ratings (all 878 pairs) and mean scores; higher $r$ per row in bold; per-style 95\% intervals are about $\pm0.2$. $^{\dagger}$Unseen styles.}
\label{tab:per_style}
\small
\begin{tabular}{@{}lccccc@{}}
\toprule
 & & \multicolumn{2}{c}{\textbf{Mean score}} & \multicolumn{2}{c}{\textbf{Pearson $r$}} \\
\cmidrule(lr){3-4}\cmidrule(l){5-6}
\textbf{Style} & $n$ & Human & Student & Student & Teacher \\
\midrule
Abstract & 70 & 5.2 & 4.3 & 0.34 & \textbf{0.40} \\
Charcoal & 74 & 4.8 & 6.7 & \textbf{0.30} & 0.23 \\
Cubist & 74 & 5.3 & 5.1 & \textbf{0.57} & 0.45 \\
Expressive & 75 & 5.6 & 6.3 & \textbf{0.51} & 0.42 \\
Gouache & 70 & 6.1 & 6.7 & \textbf{0.45} & 0.38 \\
Impressionist & 74 & 5.1 & 7.0 & \textbf{0.51} & 0.40 \\
Minimalist & 73 & 6.4 & 3.3 & \textbf{0.39} & 0.15 \\
Mosaic & 74 & 5.1 & 4.4 & \textbf{0.14} & 0.12 \\
Pencil & 76 & 4.8 & 7.0 & 0.26 & \textbf{0.49} \\
Watercolor & 68 & 5.5 & 7.8 & \textbf{0.29} & 0.27 \\
\midrule
Acrylic$^{\dagger}$ & 75 & 5.7 & 6.0 & \textbf{0.45} & 0.44 \\
Pop Art$^{\dagger}$ & 75 & 5.3 & 2.7 & \textbf{0.40} & 0.38 \\
\bottomrule
\end{tabular}
\end{table}

\begin{table}[t]
\caption{Correlation with human ratings on the 728 seen-style pairs of \S\ref{sec:setup}. Distances (Gram, LPIPS) are negated, so higher is better in every row. }
\label{tab:human_correlation}
\begin{tabular}{@{}lcc@{}}
\toprule
\textbf{Metric} & \textbf{Pearson $r$} & \textbf{Spearman $\rho$} \\
\midrule
Gram loss (negated)~\cite{gatys2016image} & $\phantom{-}0.041$ & $\phantom{-}0.049$ \\
LPIPS (negated)~\cite{zhang2018unreasonable} & $-0.404$ & $-0.394$ \\
CLIP similarity~\cite{radford2021learning}         & $\phantom{-}0.168$ & $\phantom{-}0.178$ \\
\midrule
Untuned base (0.8B, zero-shot)$^{\ddagger}$ & $\phantom{-}0.019$ & $\phantom{-}0.017$ \\

\modelname{} (ours, SFT, 0.8B)                & $\phantom{-}\mathbf{0.291}$ & $\phantom{-}\mathbf{0.325}$ \\
\modelname{} (ours, teacher, 72B)             & $\phantom{-}0.327$ & $\phantom{-}0.337$ \\
\midrule
Human inter-rater agreement$^{\dagger}$       & $\phantom{-}0.250$ & --- \\
\bottomrule
\end{tabular}

\vspace{0.2cm}
{\footnotesize $^{\dagger}$Measured on a separate $n{=}50$ paired pilot (Table~\ref{tab:zeroshot}); indicative only (95\% CI $[-0.03,0.49]$ includes zero), not a ceiling; different sample, partly different raters. $^{\ddagger}$Scored on the 711 of 728 pairs for which it returned a parseable score.}
\end{table}

\subsection{Teacher-Imitation Accuracy}
\label{sec:teacher_accuracy}

As a secondary, internal-consistency check, we measure exact-match agreement between
\modelname{} and its 72B teacher on 800 samples from the two validation styles, expressive and
mosaic, which are held out from training (Table~\ref{tab:k_ablation}: $K=0$ is the untuned base model and
$K=2$ the \modelname{} student). This complements, but does not replace, the human-correlation result. The SFT adapter reaches 28.2\% exact-match agreement with the teacher, a
25.6$\times$ improvement over the untuned 1.1\% base rate, while reducing the average length of its generated reasoning-and-score response
from 8{,}088 to 2{,}606 characters ($-67.8\%$).  Agreement within one score point rises from 12.4\% to 54.4\%, and the mean absolute error falls from 5.18 to 2.11. On a single NVIDIA T4, the student scores a pair in 39\,s (median over the 900 sampled pairs: the 878 rated ones and 22 unrated ones, \S\ref{sec:setup}) with 2.5\,GB of peak GPU memory.

The teacher-agreement result above is measured on styles absent from training, and the human-correlation result on the 150 unseen-style pairs ($r=0.381$) is comparable to that on seen styles. This is consistent with \modelname{} having learned a generalisable evaluation procedure, but caveats apply: the validation styles were also used to choose $K$ and $\lambda$, the unseen-style human sample is small, and the prompt still supplies exemplars and text style markers that carry part of the style-specific calibration.

\subsection{Qualitative Analysis}
\label{sec:qualitative}

Table~\ref{tab:qualitative} contrasts base and SFT predictions against the teacher score,
which is a 72B score and not a human rating, for four illustrative samples: rows 1--2 are one point off the teacher and rows 3--4 match it. Row 3 shows the
base model badly overconfident (9 vs.\ a teacher score of 3), which SFT correctly flags as a
failure. Row 4 falls in the clamp regime of Eq.~\ref{eq:final_score} (teacher
CP$=2\le3 \Rightarrow S_{\text{final}}=\min(\mathrm{RQ},\mathrm{CP})=2$): SFT's raw scores
($2/9/6$) do not themselves satisfy the clamp condition, yet its composite still matches the
teacher's. Across the 26 teacher-flagged clamp cases in the 800 validation samples, SFT's raw scores land in the
clamp regime themselves in only 4, suggesting it under-recognises clamp conditions even when
its final score happens to agree, a gap we return to in \S\ref{sec:discussion}.

\begin{table}[t]
\caption{Base vs.\ SFT predictions (SF/CP/RQ $\to$ composite by Eq.~\ref{eq:final_score}); m$k$ is morph frame $k$. \textbf{Teacher} is the 72B score, \emph{not} a human rating.}
\label{tab:qualitative}
{\small
\begin{tabular}{@{}p{1.4cm} p{1.85cm} p{1.85cm} c c@{}}
\toprule
\textbf{Style} &
\textbf{Base} &
\textbf{SFT} &
\textbf{Teacher} &
\textbf{Match?} \\
\midrule
Acrylic (m8)
  & 8/9/8 ${\to}8$
  & 7/8/8 ${\to}7$
  & 8
  & $\times$ \\
Pop Art (m4)
  & 8/8/8 ${\to}8$
  & 5/7/6 ${\to}5$
  & 6
  & $\times$ \\
Acrylic (m2)
  & 9/9/8 ${\to}9$
  & 3/8/7 ${\to}3$
  & 3
  & \checkmark \\
Expressive (m1)
  & 10/10/10 ${\to}10$
  & 2/9/6 ${\to}2$
  & 2
  & \checkmark \\
\bottomrule
\end{tabular}}
\end{table}

\begin{figure*}[t]
  \centering
  \includegraphics[width=\textwidth]{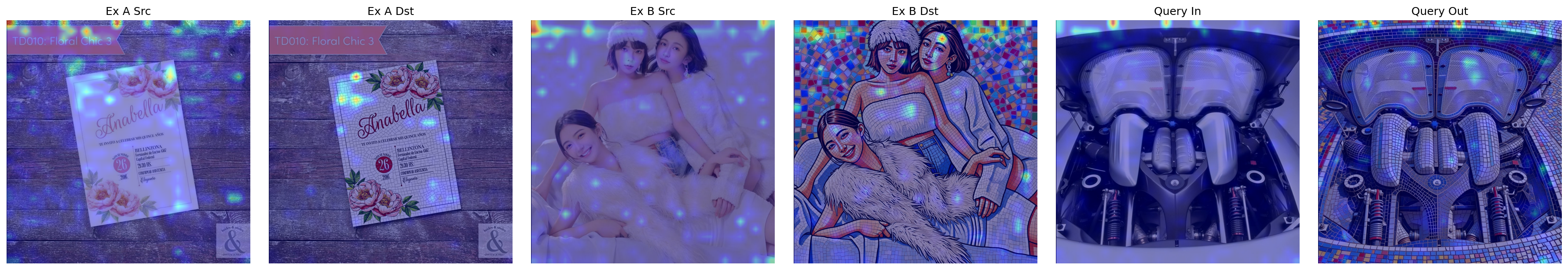}
  \caption{Attention rollout for a mosaic query given two exemplar pairs (Ex A, Ex B).}
  \Description{Six-panel figure showing two in-context exemplar pairs and a query source/output
  pair for a mosaic style transfer, overlaid with attention rollout heatmaps.}
  \label{fig:attention_rollout}
\end{figure*}

Attention rollout (Figure~\ref{fig:attention_rollout}; additional examples in the project repository~\cite{stylejudgecode2026}) concentrates on prompt-targeted style markers, \eg, tile-and-grout structure for
mosaic, alongside low-intensity background attention consistent with a ViT ``attention sink''
artefact. These maps are illustrative and are not causal evidence about which cues the model uses.

\subsection{Ablation Studies}
\label{sec:ablation}

The results above rest on two design choices fixed in \S\ref{sec:method}: two in-context
exemplar pairs ($K=2$) and a score-digit loss weight of $\lambda=5.0$. We now test both
directly.

\subsubsection{Number of In-Context Exemplars ($K$)}
\label{sec:k_ablation}

We validate the choice of $K=2$ with a controlled sweep over $K \in \{0,1,2,3\}$ on 800 samples from the two validation styles (expressive, mosaic), measuring exact-match accuracy, accuracy within $\pm1$ score
point, mean absolute error (MAE) of the predicted composite score against the teacher score,
and average output length. Table~\ref{tab:k_ablation} summarises the results.

\begin{table}[t]
\caption{Ablation on the number of exemplar pairs $K$ (the same 800 validation-style items for every $K$, all scored against the teacher's labels in the $K{=}3$ annotation set). $K{=}0$ is the untuned base model, $K{=}2$ the \modelname{} student; each $K{\ge}1$ has its own adapter.}
\label{tab:k_ablation}
\begin{tabular}{@{}ccccc@{}}
\toprule
$K$ & MAE $\downarrow$ & Exact (\%) & $\pm1$ (\%) & Avg.\ Length \\
\midrule
0 & 5.179 & 1.1  & 12.4 & 8{,}088 chars \\
1 & 2.842 & 22.8 & 43.1 & 2{,}689 chars \\
2 & \textbf{2.109} & \textbf{28.2} & \textbf{54.4} & 2{,}606 chars \\
3 & 2.600 & 19.6 & 44.0 & 3{,}527 chars \\
\bottomrule
\end{tabular}
\end{table}

Moving from $K=0$, the fully zero-shot setting, to $K=1$ yields the largest single jump in
accuracy (1.1\%$\to$22.8\% exact match), although this jump combines fine-tuning with exemplar conditioning, because $K=0$ is the untuned base model. $K=2$ gives a further, smaller gain (28.2\% exact, 54.4\% within
$\pm1$, lowest MAE at 2.109). Beyond this point, $K=3$ degrades performance on every metric
(exact match falls to 19.6\%, MAE rises to 2.600) despite the longer, more example-rich
prompt. Because each adapter is trained and evaluated with the same $K$, this drop is not a train/test mismatch. With 512-pixel images, a third pair adds only several hundred tokens, so the 8{,}192-token budget is unlikely to be the main cause; the $K=3$ adapter was, however, trained on fewer examples (3{,}860 vs.\ 5{,}614) and writes longer outputs (2{,}606$\to$3{,}527 characters), so the sweep does not isolate $K$ alone.

\subsubsection{Score-Weighted Loss}
\label{sec:loss_ablation}

 The weight $\lambda=5.0$ was
selected via a brute-force sweep over $\lambda \in \{1,2,3,4,5\}$ on the validation set, with
$\lambda=5.0$ giving the best balance between score accuracy and reasoning coherence; smaller
values reproduced the miscalibration described in \S\ref{sec:distillation}, where the model
generated fluent reasoning but poorly correlated terminal scores. The sweep stops at $\lambda=5.0$, the largest value tried, so a larger weight may do better.

\subsection{Discussion and Limitations}
\label{sec:discussion}

\modelname{} is best read as a much cheaper, more human-aligned substitute for a human panel, not a perfect replica of one: style judgment is noisy, and with single-rated pairs we cannot establish a human ceiling or claim human parity. Reference pairs give the model an operational definition of the target style, which matters most where the transformation is visually non-obvious. Exact-match teacher agreement is a strict criterion, and the 0.8B student cannot replicate the full depth of a 72B teacher's reasoning, particularly for partially stylised frames. The RQ/CP clamp threshold of Eq.~\ref{eq:final_score} was set heuristically, and the student recognises the clamp regime in its own raw scores in only 4 of 26 flagged cases (\S\ref{sec:qualitative}). The score-digit weight of Eq.~\ref{eq:loss} is a static heuristic; auxiliary regression heads or reward-model objectives (\eg, DPO) are more principled alternatives that we did not test.

Further limitations: the 728-pair benchmark covers seen styles and the unseen-style sample has 150 pairs; the three raters are project-team members without art training; prompts include text style markers, so conditioning is not exemplar-only; all images come from one generator family; $\lambda$ was swept only over 1--5, without an unweighted ($\lambda{=}0$) run;  the model's written rationales were not compared with the raters' free-text justifications; and we did not run CSD~\cite{somepalli2024csd} or ArtFID~\cite{wright2022artfid}. LPIPS measures content preservation, so its negative correlation reflects a benchmark that rewards stylization rather than a flaw in LPIPS; an unsigned stylization-strength proxy such as raw LPIPS distance ($r=0.405$) is therefore a strong baseline on this benchmark, because most queries come from graded morph strips, and a naturally sampled set would separate the two more fairly. Gram statistics show essentially no correlation. Finally, because benchmark items were sampled uniformly over the teacher's score, their quality range is wider than in a naturally drawn sample; this can raise all correlations and may favour the teacher, so results on a naturally sampled set remain to be established.

\textbf{Future work.} A multi-rater benchmark would give a same-set human ceiling, other generators' outputs would test cross-generator transfer, and per-style calibration (\S\ref{sec:main_results}) could remove the style-level offsets; \modelname{} could also serve as a reward signal for training stylization models (an exploratory GRPO notebook in the repository is not part of the reported results).

\section{Conclusion}
\label{sec:conclusion}

We presented \modelname{}, an in-context learning framework for evaluating artistic style transfer, distilled from a 72B teacher into a 0.8B student that runs on a single 16\,GB GPU (NVIDIA T4), and validated it directly against 878 human ratings. It correlates with human judgment at $r=0.291$ on seen and $0.381$ on unseen styles, the strongest positive correlation among the metrics we tested on this benchmark (LPIPS correlates negatively), and agrees with its teacher on 28.2\% of held-out validation items with 67.8\% shorter outputs. Per style, its correlation with human ratings has a positive point estimate in all twelve styles, and most of its remaining error is style-level calibration rather than within-style ranking. We release the human-rating benchmark and guideline, our SDXL-adapted FreeMorph implementation, and code and weights for local inference~\cite{stylejudgecode2026}, so that future metrics can be scored on the same pairs.

\begin{acks}
We thank the Samsung PRISM program for its support. AI tools (large language models) assisted with writing, code and figure preparation; the authors verified all results and take responsibility for the content.
\end{acks}

\bibliographystyle{ACM-Reference-Format}
\bibliography{ICVGIP-Latex-Template}

\noindent\textbf{Supplementary material:} \url{https://github.com/tomoeOOseven/StyleJudge}

\end{document}